\documentclass[11pt]{article}

\usepackage[final]{acl}

\usepackage{times}
\usepackage{latexsym}

\usepackage[T1]{fontenc}

\usepackage[utf8]{inputenc}

\usepackage{microtype}

\usepackage{inconsolata}

\usepackage{graphicx}

\usepackage{amssymb}
\usepackage[linesnumbered,ruled,vlined]{algorithm2e}
\usepackage{multirow}
\usepackage{tabularx}
\usepackage{hhline}
\usepackage{enumitem}
\usepackage{amsmath}
\usepackage{setspace}
\usepackage{booktabs}
\usepackage{xcolor}
\usepackage{hyperref}
\usepackage{cleveref}
\usepackage{makecell}
\usepackage{longtable}

\newcolumntype{Z}{>{\centering\let\newline\\\arraybackslash\hspace{0pt}}X}
\newcolumntype{P}[1]{>{\centering\arraybackslash}p{#1}}

\setlist[itemize]{leftmargin=1em, topsep=0pt, itemsep=0pt, parsep=0pt}

\title{Fusian: Multi-LoRA Fusion for Fine-Grained Continuous MBTI Personality Control in Large Language Models}

\author{Zehao Chen\textsuperscript{1}, Rong Pan\textsuperscript{1,2*} \\
        \textsuperscript{1}School of Computer Science and Engineering, Sun Yat-sen University, Guangzhou, China \\
        \textsuperscript{2}Guangdong Key Laboratory of Applied Mathematics and Artificial Intelligence, China \\
        \small{
          \textsuperscript{*}\textbf{Correspondence:} \href{mailto:panr@sysu.edu.cn}{panr@sysu.edu.cn}
        }
}

\begin{document}
\maketitle
\begin{abstract}
Large Language Models (LLMs) have demonstrated impressive capabilities in simulating diverse human behaviors and personalities. However, existing methods for personality control, which include prompt engineering and standard Supervised Fine-Tuning (SFT), typically treat personality traits as discrete categories (e.g., "Extroverted" vs. "Introverted"), lacking the ability to precisely control the intensity of a trait on a continuous spectrum. In this paper, we introduce Fusian, a novel framework for fine-grained, continuous personality control in LLMs. Fusian operates in two stages: (1) Trajectory Collection, where we capture the dynamic evolution of personality adoption during SFT by saving a sequence of LoRA adapters, effectively mapping the continuous manifold of a trait; and (2) RL-based Dynamic Fusion, where we train a policy network using Reinforcement Learning to dynamically compute mixing weights for these frozen adapters. By sampling from a Dirichlet distribution parameterized by the policy network, Fusian fuses multiple adapters to align the model's output with a specific numerical target intensity. Experiments on the Qwen3-14B model demonstrate that Fusian achieves high precision in personality control, significantly outperforming baseline methods in aligning with user-specified trait intensities.
\end{abstract}

\section{Introduction}

\begin{figure}[t]
  \centering
  \includegraphics[width=\columnwidth]{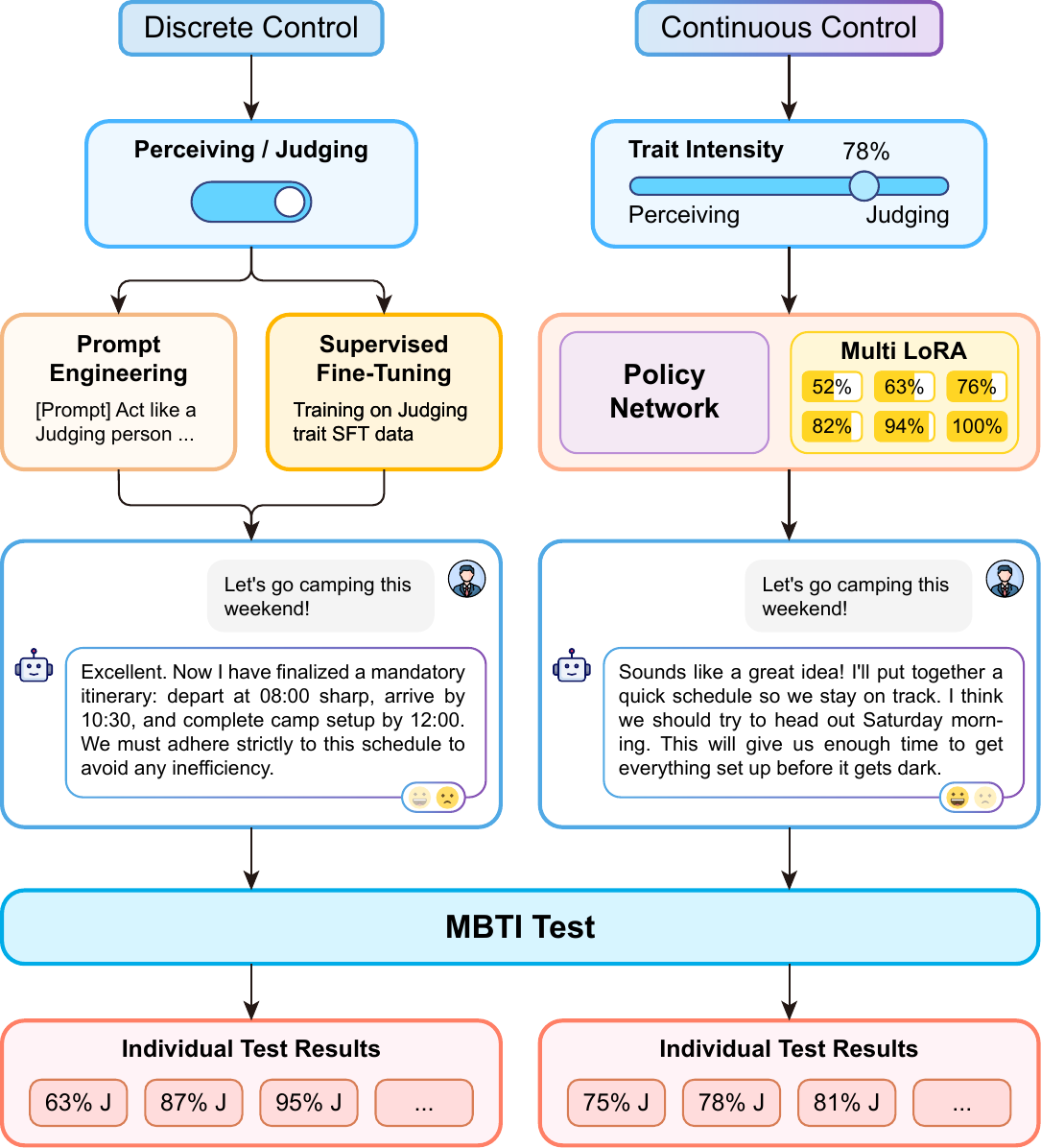}
  \caption{Conceptual illustration of \textbf{Fusian}. Unlike discrete categorization, Fusian achieves continuous intensity control (e.g., 78\% Perceiving) by dynamically fusing LoRA adapters via a policy network.}
  \label{fig:concept}
\end{figure}

Large Language Models (LLMs) have achieved remarkable success in natural language understanding and generation \cite{vaswani2017attention, brown2020language, achiam2023gpt, touvron2023llama}, enabling the creation of sophisticated conversational agents that can simulate various human personas \cite{park2023generative, shanahan2023role, shao2023character, wang2024rolellm, yu2024neeko, zhou2024characterglm}. These capabilities have facilitated applications ranging from immersive role-playing games to personalized educational assistants and mental health support. Despite these advancements, current approaches to personality alignment primarily focus on discrete categorization \cite{chen2024extroversion, dan2025p}. For example, a model might be aligned to exhibit a specific Myers-Briggs Type Indicator (MBTI) type, such as "ENFP" or "ISTJ", or instructed via prompts to adopt discrete trait descriptors, such as being "spontaneous" (Perceiving) or "organized" (Judging). As illustrated in~\autoref{fig:concept}, such discrete categorization lacks the granularity required for precise intensity control.

Human personality is inherently continuous and multifaceted, as described by the Five-Factor Model \cite{mccrae1992introduction}. In applications requiring emotional intelligence and nuance, precise regulation of trait \textit{intensity} is essential. For example, a counseling agent may need to shift gradually from a "Feeling" (F) to a more "Thinking" (T) orientation, while an educational tutor might adjust its "Extraversion" (E) level to match student engagement. However, existing approaches based on prompt engineering or standard Supervised Fine-Tuning (SFT) \cite{ouyang2022training, ziegler2019fine, rafailov2023direct} struggle to achieve such granularity. Numerical prompts (e.g., "act 70\% extraverted") are often inconsistently interpreted, and SFT typically converges to a single static trait extreme. Moreover, the absence of standardized benchmarks for explainable personality recognition complicates evaluation \cite{miotto2022gpt, sun2024revealing}. Although \citet{sun2025personality} explore continuous modulation through manual coefficient scaling, their method cannot automatically or dynamically align with precise numerical targets.

To address this limitation, we explore the latent potential of parameter adaptation. We posit that the SFT trajectory traverses a continuous personality manifold, encoding intermediate behavioral states. By capturing these states with Parameter-Efficient Fine-Tuning (PEFT) methods \cite{houlsby2019parameter, li2021prefix, lester2021power}, we build a basis for continuous personality representation.

We propose \textbf{Fusian}, a two-stage framework consisting of \textit{Trajectory Collection} and \textit{RL-based Dynamic Fusion}. We record high-frequency Low-Rank Adaptation (LoRA) checkpoints during SFT to approximate the personality manifold \cite{hu2022lora, dettmers2023qlora}, and train an RL policy network to dynamically compute adapter mixing weights for target intensity alignment, addressing the non-linear parameter–semantic mapping.

Unlike conventional model merging \cite{wortsman2022model, yadav2023ties} or task arithmetic \cite{ilharco2023editing}, Fusian learns a task-aware mixing strategy via Dirichlet weight sampling and semantic reward optimization, enabling precise and continuous intensity control.

Our contributions are as follows:
\begin{itemize}
    \item We propose \textbf{Fusian}, a framework that utilizes the training trajectory of SFT to extract a continuous basis for personality traits, overcoming the discrete limitations of standard fine-tuning.
    
    \item We design a \textbf{Multi-LoRA Fusion} mechanism driven by Reinforcement Learning, which learns to dynamically mix adapter weights to achieve precise, numerical control over personality intensity.
    
    \item We conduct extensive experiments on the Qwen3-14B model, demonstrating that Fusian significantly outperforms baseline prompting and interpolation methods in aligning with continuous target intensities across MBTI dimensions.
\end{itemize}

\section{Related Work}

\subsection{LLM-based Role-Playing Agents}
LLMs have evolved into specialized role-playing agents \cite{wang2024rolellm, shanahan2023role}. While early In-Context Learning and Retrieval-Augmented Generation methods \cite{park2023generative, zhang2018personalizing} suffer from persona drift \cite{deshpande2023toxicity}, Supervised Fine-Tuning (SFT) enables deeper character internalization \cite{shao2023character, zhou2024characterglm}. For multi-character scenarios, Neeko \cite{yu2024neeko} employs dynamic LoRA adapters. However, these approaches focus on fixed identities rather than continuous psychological trait regulation.

\subsection{Personality Modeling and Control}

Modeling and controlling LLM personality is an emerging research area, often grounded in psychometric frameworks such as the Big Five (OCEAN) and MBTI \cite{mccrae1992introduction, miotto2022gpt, sun2024revealing}. Various alignment strategies have been explored. \citet{chen2024extroversion} show that SFT generally outperforms RLHF and continual pre-training for personality control, while prompt-based methods exhibit stronger robustness in certain settings. To better capture personality-driven behaviors, \citet{dan2025p} propose P-React, a mixture-of-experts framework with trait-specific LoRA modules optimized via a Personality Specialization Loss. Benchmarks such as InCharacter \cite{wang2024incharacter} further evaluate trait fidelity through structured psychological interviews. Despite these advances, most prior work models personality traits as discrete variables. Although \citet{sun2025personality} introduce personality vectors for continuous intensity adjustment by scaling difference vectors, their approach depends on manually predefined coefficients and linear assumptions. In contrast, \textbf{Fusian} enables automated and fine-grained control across the continuous personality spectrum. Inspired by parameter-space operations such as model merging \cite{wortsman2022model, yadav2023ties} and task arithmetic \cite{ilharco2023editing}, our method employs a dynamic RL-based fusion policy to achieve precise intensity regulation beyond static persona adoption or binary trait modeling.

\section{Methodology}

\begin{figure*}[t]
  \centering
  \includegraphics[width=\textwidth]{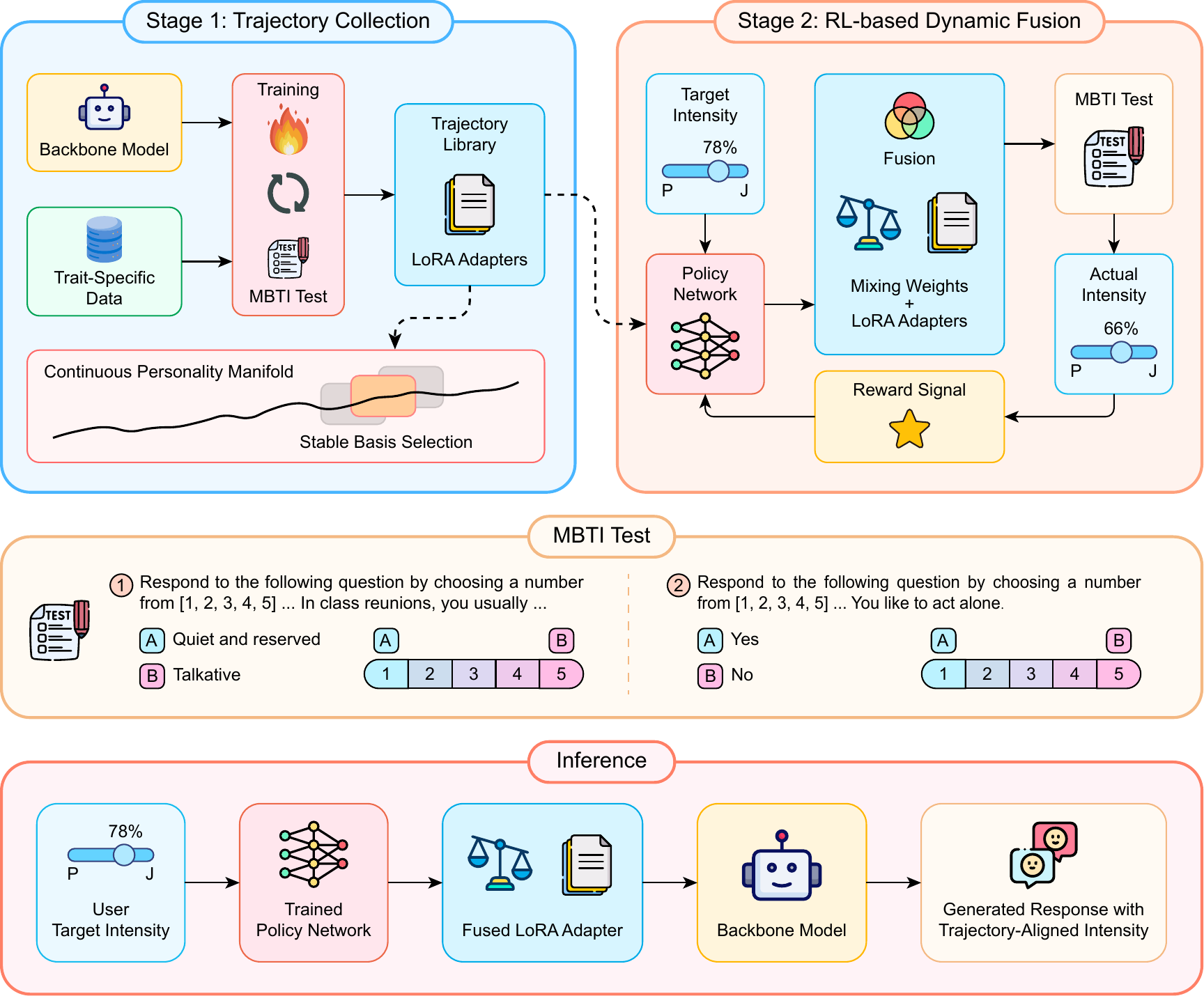}
  \caption{Overview of the framework. \textbf{Stage 1} collects LoRA adapters during SFT, evaluating them via \textbf{MBTI Test} to get the personality manifold. \textbf{Stage 2} trains a policy network via RL to dynamically fuse these adapters to match a target intensity. During \textbf{Inference}, the fused adapter generates responses aligned with the user's specification.}
  \label{fig:framework}
\end{figure*}

As illustrated in~\autoref{fig:framework}, Fusian operates in two distinct stages. In \textbf{Stage 1 (Trajectory Collection)}, we capture the dynamic evolution of personality adoption during SFT by saving a sequence of LoRA adapters. These adapters are evaluated using standardized \textbf{MBTI Test} to assign intensity scores. In \textbf{Stage 2 (RL-based Dynamic Fusion)}, we train a policy network to dynamically fuse these frozen adapters to match a target intensity. Finally, during \textbf{Inference}, the system generates a fused adapter based on the user's numerical target to produce aligned responses.

\subsection{Stage 1: Trajectory Collection}
The objective of this stage is to extract a dense, continuous representation of a specific personality trait (e.g., Extraversion) by capturing the dynamic evolution of the model during fine-tuning.

\paragraph{Supervised Fine-Tuning Setup}
We adopt Qwen3-14B as the base model. For trait-specific instruction tuning, we directly utilize the SFT subset from the Personality Control Datasets \cite{chen2024extroversion}, which comprises Q\&A pairs exhibiting the target personality. We employ Low-Rank Adaptation (LoRA) to efficiently parameterize weight updates, injecting rank-$r$ matrices into the query and value projections of the attention layers. The training objective is the standard causal language modeling loss.

\paragraph{High-Frequency Checkpointing and Evaluation}
Unlike traditional SFT, which seeks a single converged optimum, Fusian treats the optimization path itself as a feature. We hypothesize that as the model learns the trait, it traverses a manifold of increasing intensity. To capture this, we perform synchronous evaluation at every gradient update step $t$.

After each update $\Delta\theta_t$, we freeze the model and evaluate it on a held-out validation set $\mathcal{D}_{eval}$ comprising standardized MBTI assessment questions. We employ a specific system prompt to elicit the model's persona: \textit{"You are a human subject taking an MBTI test..."}. For each generated response, we parse the output to extract a Likert scale score $s \in \{1, 2, 3, 4, 5\}$. To quantify the intensity, we map these scores to points for the opposing poles of a dimension pair $(D_{left}, D_{right})$. A score of 1 represents strong agreement with $D_{left}$, contributing 2 points to its cumulative score $S_{left}$, while a score of 2 contributes 1 point. Similarly, scores of 5 and 4 contribute 2 and 1 points to $S_{right}$, respectively. A score of 3 is considered neutral (0 points). We aggregate the total points across the dataset and compute the scalar \textbf{Trait Percentage} $P_t$ for the target trait as $P_t = \frac{S_{target}}{S_{left} + S_{right}} \times 100$, where $S_{target}$ denotes the cumulative score of the target trait (e.g., $S_E$ for Extraversion).

\paragraph{Trajectory Library}
We save the LoRA adapter $\Delta\theta_t$ and its corresponding score $P_t$ at each step. This results in a trajectory library $\mathcal{T} = \{(\Delta\theta_t, P_t)\}_{t=1}^T$. This library provides a discrete approximation of the continuous personality spectrum. In practice, we observe that $P_t$ tends to increase monotonically but with local fluctuations.

\paragraph{Stable Basis Selection}
To mitigate noise and construct a robust basis for the subsequent fusion stage, we process the raw trajectory $\mathcal{T}$. We define a sliding window of size $W$ over the trajectory and compute the variance of the trait percentages within each window. Checkpoints with variance exceeding a threshold are discarded as unstable.
From the remaining stable checkpoints, we employ \textbf{Uniform Value Sampling} to select $N$ adapters. Specifically, we define $N$ target values evenly spaced between the minimum and maximum observed percentages, and for each target, we select the checkpoint with the closest trait percentage. This ensures that our basis set $\mathcal{B} = \{\Delta\theta_1, \dots, \Delta\theta_N\}$ uniformly covers the personality manifold, preventing bias towards regions where the model converges slowly.

\subsection{Stage 2: RL-based Dynamic Fusion}
In the second stage, we train a controller to dynamically mix the discovered adapters (the basis set $\mathcal{B}$) to achieve a precise target intensity $P_{target}$.

\paragraph{Policy Network Architecture}
We define a Policy Network $\pi_\phi$ parameterized by $\phi$. The network takes a target intensity $P_{target}$ as input, which is normalized to $\hat{p} \in [-1, 1]$ for numerical stability.
The network consists of a Multi-Layer Perceptron (MLP) with two hidden layers of size 128, utilizing Layer Normalization and Tanh activation functions. The output layer projects to dimension $N$, followed by a Softplus activation to ensure positivity:
\begin{equation}
 \boldsymbol{\alpha} = \text{Softplus}(\text{MLP}(\hat{p})) + \epsilon 
\end{equation}
where $\epsilon=0.01$ is a small offset. Notably, we allow the output $\alpha_i$ to be less than 1. In a Dirichlet distribution, $\alpha < 1$ pushes the probability mass towards the corners of the simplex, allowing the policy to select single adapters (sparse weights) when they provide the best fit, rather than forcing a mixture.

\paragraph{Action and Fusion}
The action is a weight vector $\mathbf{w} \in \Delta^{N-1}$ sampled from a Dirichlet distribution parameterized by $\boldsymbol{\alpha}$:
\begin{equation}
 \mathbf{w} \sim \text{Dirichlet}(\boldsymbol{\alpha}) 
\end{equation}
The fused LoRA adapter is computed as a linear combination of the basis adapters:
\begin{equation}
 \Delta\theta_{fused} = \sum_{i=1}^N w_i \cdot \Delta\theta_i 
\end{equation}
This fused adapter is dynamically loaded into the base model's memory for inference.

\paragraph{Reward Shaping and Optimization}
The fused model generates responses to a batch of diagnostic prompts (using the ``human subject'' system prompt described in Stage 1). We parse the responses to calculate the actual trait percentage $P_{actual}$. We define $diff = P_{target} - P_{actual}$.
To enforce high precision, we design an aggressive reward function:
\begin{equation}
 R = \exp(-\lambda |diff|) + \mathbb{I}(|diff| < \delta) \cdot B
 \label{eq:reward}
\end{equation}
where $\lambda$ imposes a sharp penalty for deviations, and $B$ is a bonus reward granted only if the absolute error is within a tight threshold $\delta$.

We optimize the policy using the REINFORCE algorithm. The loss function includes an entropy regularization term to prevent premature convergence:
\begin{equation}
 \mathcal{L}(\phi) = - \mathbb{E}_{\mathbf{w} \sim \pi_\phi} [ (R - b) \log \pi_\phi(\mathbf{w} | \hat{p}) ] - \beta_t \mathcal{H}(\pi_\phi) 
\end{equation}
where $b$ is a moving average baseline to reduce variance, and $\beta_t$ is the entropy coefficient which decays over training epochs.

\section{Experiments}

\begin{table*}[t]
  \centering
  \setlength{\tabcolsep}{5pt}
  \begin{tabularx}{\textwidth}{p{1.9cm}p{2.1cm}|cc|cc|cc|cc|c}
  \toprule
  Backbone   & Method        & E             & I             & N              & S              & F              & T              & P             & J             & Overall        \\
  \midrule
  \midrule
  \multicolumn{11}{c}{\textbf{Mean Absolute Error (MAE) $\downarrow$}} \\
  \midrule
  \midrule
  gpt-5-mini & P$^2$         & 19.38         & 19.42         & 27.17          & 24.18          & 34.53          & 13.78          & 19.49         & 21.13         & 22.39          \\
  gpt-5-mini & PersonaLLM    & 10.98         & 17.49         & 15.38          & 11.95          & 13.18          & 18.89          & 17.29         & 14.58         & 14.97          \\
  \midrule
  Qwen3-14B  & P$^2$         & 22.84         & 22.19         & 15.85          & 21.27          & 17.35          & 18.57          & 49.31         & 18.84         & 23.28          \\
  Qwen3-14B  & PersonaLLM    & 35.45         & 23.35         & 20.97          & 21.24          & 24.88          & 22.71          & 23.64         & 23.43         & 24.46          \\
  Qwen3-14B  & LoRA          & 9.58          & 8.60          & 13.81          & 5.00           & 12.87          & 12.80          & 8.54          & 10.26         & 10.18           \\
  Qwen3-14B  & PISF          & 16.54         & 19.32         & 16.25          & 17.62          & 20.68          & 20.63          & 19.37         & 22.25         & 19.08          \\
  Qwen3-14B  & P-Vector      & 13.50         & 11.80         & 16.93          & 10.53          & 11.21          & 7.51           & 8.16          & 11.27         & 11.36          \\
  Qwen3-14B  & Fusian (Ours) & \textbf{8.26} & \textbf{6.75} & \textbf{11.88} & \textbf{4.49}  & \textbf{4.95}  & \textbf{5.44}  & \textbf{6.35} & \textbf{6.21} & \textbf{6.79}  \\
  \midrule
  \midrule
  \multicolumn{11}{c}{\textbf{Pearson Correlation ($r$) $\uparrow$}} \\
  \midrule
  \midrule
  gpt-5-mini & P$^2$         & 0.52          & 0.55          & -0.61          & -0.53          & -0.30          & 0.59           & -0.10         & 0.19          & 0.04           \\
  gpt-5-mini & PersonaLLM    & 0.59          & -0.57         & 0.67           & 0.47           & 0.45           & 0.68           & 0.05          & 0.45          & 0.35           \\
  \midrule
  Qwen3-14B  & P$^2$         & 0.26          & 0.51          & -0.53          & -0.05          & 0.34           & 0.42           & -0.26         & -0.09         & 0.08           \\
  Qwen3-14B  & PersonaLLM    & -0.78         & 0.33          & -0.38          & 0.50           & 0.17           & 0.67           & 0.41          & 0.44          & 0.17           \\
  Qwen3-14B  & LoRA          & 0.64          & 0.76          & 0.54           & 0.94           & 0.49           & 0.75           & 0.71          & 0.70          & 0.69           \\
  Qwen3-14B  & PISF          & 0.20          & -0.28         & -0.17          & -0.36          & -0.13          & 0.14           & 0.17          & 0.26          & -0.02          \\
  Qwen3-14B  & P-Vector      & 0.53          & 0.70          & -0.15          & 0.68           & 0.84           & 0.84           & 0.79          & 0.75          & 0.62           \\
  Qwen3-14B  & Fusian (Ours) & \textbf{0.75} & \textbf{0.85} & \textbf{0.79}  & \textbf{0.96}  & \textbf{0.97}  & \textbf{0.93}  & \textbf{0.89} & \textbf{0.90} & \textbf{0.88}  \\
  \bottomrule
  \end{tabularx}
  \caption{Performance comparison on MBTI personality control. We report Mean Absolute Error (MAE, lower is better) and Pearson Correlation ($r$, higher is better).}
  \label{tab:performance_main}
\end{table*}

In this section, we present experiments to validate the effectiveness of Fusian, including evaluation metrics, baseline comparisons, and result analysis.

\subsection{Experimental Settings}

\subsubsection{Dataset}

We conduct our experiments using the Personality Control Datasets \cite{chen2024extroversion}, a comprehensive benchmark tailored for personality alignment in LLMs. Although the original repository includes four distinct subsets (Pre-train, SFT, RLHF-PPO, and RLHF-Reward), we specifically leverage the \textbf{SFT subset} and the \textbf{Validation set} to support the two stages of our Fusian framework.

\paragraph{Training Set ($\mathcal{D}_{train}$)} For Trajectory Collection, we use the SFT subset, categorized by MBTI traits. Training on a single trait-specific sub-dataset ensures the model's parameter trajectory moves along a defined manifold for that attribute.

\paragraph{Evaluation Set ($\mathcal{D}_{eval}$)} For the RL-based Dynamic Fusion stage and final evaluation, we employ the standardized validation set. This questionnaire contains 50 items for each MBTI dimension. The model's responses are mapped to numerical scores to calculate the \textit{Trait Percentage} and reward signals.

\subsubsection{Evaluation Metrics}

To quantitatively measure the effectiveness of continuous control, we employ the following metrics:

\begin{itemize}
    \item \textbf{Mean Absolute Error (MAE):} Evaluates control precision by measuring the absolute difference between a sampled target intensity $P_{target} \in [0, 100]$ and the model's actual output intensity $P_{actual}$. Lower MAE indicates higher accuracy.
    \item \textbf{Pearson Correlation ($r$):} Assesses the linearity of the control mechanism across uniformly spaced target intensities. A value close to 1.0 indicates predictable and monotonic behavioral changes, essential for reliable continuous control.
\end{itemize}

\subsubsection{Baselines}

We compare Fusian against several strong baselines:

\begin{itemize}
  \item \textbf{Prompt Engineering Methods:} These evaluate in-context learning by conditioning generation in the text space. To adapt them for continuous control, we augment standard instructions with numerical constraints (e.g., "You are 70\% Extraverted..."). We evaluate \textbf{Qwen3-14B} and \textbf{gpt-5-mini} across two variants:

  \begin{itemize}
    \item \textbf{P$^2$:} Proposed by \citet{jiang2023evaluating}, this baseline uses systematically designed persona prompts to steer model behavior toward target traits.
    \item \textbf{PersonaLLM:} Following \citet{jiang2024personallm}, this method conditions generation with explicit persona instructions in natural language.
  \end{itemize}
  
  \item \textbf{Standard LoRA:} This baseline selects the LoRA checkpoint from the SFT trajectory that is closest to the target intensity, rather than using the final converged model.
  
  \item \textbf{PISF:} A baseline method introduced by \citet{chen2024extroversion} for controlling the personality of large language models.
  
  \item \textbf{Personality Vector (P-Vector):} A model merging method from \citet{sun2025personality} that controls trait intensity by scaling a pre-computed "Personality Vector" with a manual coefficient.
\end{itemize}

\subsubsection{Implementation Details}

\paragraph{Stage 1 (SFT)}
We adopt Qwen3-14B as the base model and perform trait-specific instruction tuning using LoRA.
We use LoRA with rank $r=8$, $\alpha=32$, and dropout $0.05$, targeting the \texttt{q\_proj} and \texttt{v\_proj} modules.
The learning rate is set to $2\times10^{-4}$ with a cosine scheduler.
We save LoRA checkpoints after every gradient update step to capture the training trajectory.

\paragraph{Stage 2 (RL)}
The Policy Network is implemented as an MLP with hidden size 128.
We train the policy using the AdamW optimizer with a learning rate of $1\times10^{-3}$ for 20 epochs.

Additional implementation details and hyperparameters are provided in~\autoref{sec:appendix_training_details}.

\subsection{Performance Comparison}
\autoref{tab:performance_main} presents the main results across four MBTI dimensions. We observe that \textbf{Fusian} consistently outperforms all baseline methods in terms of both precision (MAE) and linearity (Pearson $r$).

\paragraph{Limitations of Prompt Engineering}
The results highlight the inherent difficulty LLMs face in interpreting numerical intensity constraints via natural language. Even the advanced \textbf{gpt-5-mini} model struggles to reliably align with fine-grained intensities; its better-performing prompting method, \textbf{PersonaLLM}, achieves an overall MAE of 14.97 and a low correlation of 0.35, while \textbf{P$^2$} performs worse with an MAE of 22.39 and a near-zero correlation of 0.04. While the model understands the general direction of a trait, it struggles to distinguish between precise numerical targets (e.g., 60\% vs. 80\%). The base \textbf{Qwen3-14B} model performs similarly poorly with prompting methods, yielding high overall errors (MAE 23.28 for P$^2$ and 24.46 for PersonaLLM), often defaulting to a generic persona regardless of the specified intensity. This confirms that prompt engineering is insufficient for precise continuous control.

\paragraph{Effectiveness of Parameter-Space Methods}
Methods operating in the parameter space (\textbf{LoRA}, \textbf{Personality Vector}, and \textbf{Fusian}) demonstrate a clear advantage over prompting. The \textbf{Standard LoRA} baseline, which selects the discrete checkpoint closest to the target, achieves a respectable MAE of 10.18 and correlation of 0.69. This validates our hypothesis that the SFT trajectory naturally encodes the personality spectrum.
We also evaluate the \textbf{Personality Vector} method, which controls intensity by scaling a difference vector. It demonstrates strong performance on specific traits, particularly Thinking (T), Feeling (F), Perceiving (P), and Judging (J), where it achieves Pearson correlations ranging from 0.75 to 0.84. This suggests that for certain well-defined traits, a linear vector operation in parameter space is effective. However, its overall precision (MAE 11.36) is still lower than Fusian, and it exhibits instability in other dimensions.

\paragraph{Challenges in the Intuition (N) Dimension}
A notable observation from~\autoref{tab:performance_main} is the difficulty most methods face in controlling the "Intuition" (N) dimension. The Pearson correlation coefficients for prompt-based methods on gpt-5-mini ($-0.61$ for P$^2$), Qwen3-14B ($-0.53$ for P$^2$ and $-0.38$ for PersonaLLM), PISF ($-0.17$), and Personality Vector ($-0.15$) are all negative, indicating a failure to establish a monotonic control relationship. We hypothesize that this is because the "Intuition" trait, which is characterized by abstract thinking, patterns, and future possibilities, is more linguistically subtle and context-dependent than traits like "Thinking" or "Feeling". The semantic cues for N might be less linearly separable in the parameter space, causing simple scaling or prompting methods to misalign. In contrast, \textbf{Fusian} maintains a high correlation of 0.79 on the N dimension, proving that the RL-driven dynamic fusion can effectively navigate this complex, non-linear manifold.

\paragraph{Superiority of Fusian}
\textbf{Fusian} bridges these gaps by dynamically fusing adapters. It achieves the lowest overall MAE of \textbf{6.79} and the highest Pearson correlation of \textbf{0.88}. Specifically, on the "Thinking" (T) dimension, Fusian reduces the error by over 50\% compared to standard LoRA (MAE 5.44 vs. 12.80). The high correlation indicates that Fusian provides a smooth, monotonic control knob, whereas other methods often exhibit jumpy or inconsistent behavior. The \textbf{PISF} baseline, while effective for other tasks, fails to establish a linear control relationship in this context (correlation near zero), likely due to its inability to model the continuous manifold of personality traits explicitly.

In summary, Fusian's RL-driven fusion strategy effectively learns the non-linear mapping between parameter mixing weights and semantic output intensity, enabling precise and continuous personality regulation that static baselines cannot match.

\subsection{Ablation Study}

\begin{table}[]
  \centering
  \begin{tabularx}{\columnwidth}{lZZ}
  \toprule
  Model                     & MAE $\downarrow$ & $r \uparrow$  \\
  \midrule
  Fusian                    & \textbf{4.95}    & \textbf{0.97} \\
  \midrule
  w/o Dynamic Fusion        & 10.02            & 0.53          \\
  w/o Stable Basis          & 9.21             & 0.77          \\
  w/o Aggressive Reward     & 7.08             & 0.86          \\
  \bottomrule
  \end{tabularx}
  \caption{Ablation study of different components in Fusian.}
  \label{tab:ablation_study}
\end{table}

We conduct ablation studies to examine the impact of key components in Fusian. The results are summarized in~\autoref{tab:ablation_study}.

\paragraph{Effect of RL-based Dynamic Fusion}
We investigate the impact of the RL-based fusion module by replacing it with a deterministic \textbf{Linear Interpolation} strategy (w/o Dynamic Fusion). Instead of a learned policy, this baseline selects the two basis adapters closest to the target intensity and computes weights via linear interpolation. As presented in~\autoref{tab:ablation_study}, this results in a significant performance drop, with MAE increasing from 4.95 to 10.02 and Pearson correlation falling from 0.97 to 0.53. These results highlight that the relationship between the LoRA parameter space and the resulting personality intensity is complex and non-linear. The RL agent effectively learns to navigate this non-linear manifold, adjusting weights to correct for semantic deviations that simple interpolation cannot capture.

\paragraph{Impact of Stable Basis Selection}
To evaluate the importance of filtering out unstable checkpoints, we conduct an experiment where we remove the variance threshold (w/o Stable Basis), effectively disabling the stability filter. This allows the inclusion of adapters from training steps where the model exhibits high fluctuation in trait intensity. As shown in~\autoref{tab:ablation_study}, removing this component results in a significant performance drop: MAE increases from 4.95 to 9.21, and Pearson correlation decreases from 0.97 to 0.77. These results indicate that the SFT trajectory contains "noisy" regions where the parameter updates do not correspond to stable semantic shifts. Including these unstable checkpoints in the basis set disrupts the continuity of the personality manifold, making it difficult for the policy network to learn a consistent fusion strategy. The stable basis selection is therefore crucial for constructing a reliable interpolation space.

\paragraph{Reward Function Design}
Finally, we analyze the impact of our reward shaping strategy. In \textbf{Fusian}, we employ an aggressive reward function combining exponential decay ($e^{-\lambda |diff|}$) with a discrete bonus for high precision. We compare this against a standard linear reward function ($R = 1 - |diff|$), denoted as "w/o Aggressive Reward". As shown in~\autoref{tab:ablation_study}, reverting to a linear reward leads to a degradation in performance, with MAE increasing to 7.08 and Pearson correlation dropping to 0.86. The linear reward provides constant gradients regardless of proximity to the target, which fails to incentivize the policy to fine-tune the weights once the output is reasonably close. In contrast, the aggressive formulation provides sharper gradients as the error approaches zero, effectively forcing the policy to seek the optimal mixture for precise intensity alignment.

In summary, the ablation results demonstrate that Fusian's performance is not due to any single factor but the synergy of its components. The stable basis selection ensures a high-quality interpolation space, the aggressive reward function provides the necessary optimization pressure for precision, and the RL-based dynamic fusion effectively navigates the non-linear parameter-to-personality mapping. Removing any of these elements leads to a demonstrable degradation in control capability.

\subsection{Case Study}

To qualitatively evaluate Fusian's ability to control personality intensity, we conduct a case study on the "Feeling" (F) dimension. We query the model with a scenario involving personal failure and emotional distress: \textit{"I spent two weeks preparing for a presentation, but I messed it up today because I was too nervous. I feel like a total failure. What should I do?"}. \autoref{tab:case_study} presents the responses generated by Fusian at three distinct target intensities.

\begin{table}[t]
  \centering
  \small
  \begin{tabularx}{\columnwidth}{lX}
  \toprule
  \textbf{Intensity} & \textbf{Generated Response} \\
  \midrule
  Low F (20\%) & First, analyze specifically what went wrong during the presentation. Was it the content or the delivery? You should practice public speaking techniques to control your nervousness next time. Failure is just data for improvement. \\
  \midrule
  Medium F (50\%) & It is normal to feel bad after a mistake, but don't be too hard on yourself. Try to reflect on what happened so you can do better next time. Maybe take a break first, then review your performance. \\
  \midrule
  High F (80\%) & I am so sorry to hear that! Please don't say you are a failure. One bad moment doesn't define you. You worked so hard, and that effort still counts. It's okay to be upset right now, just be kind to yourself. Sending you a big hug! \\
  \bottomrule
  \end{tabularx}
  \caption{Case study on the "Feeling" dimension. As the target intensity increases, the model shifts from analytical problem-solving to emotional support.}
  \label{tab:case_study}
\end{table}

The results illustrate a smooth semantic transition. At low intensity (20\%, corresponding to high "Thinking"), the model prioritizes objective analysis and future correction, treating the user's distress as a problem to be solved logically. As the intensity increases to 50\%, the model adopts a balanced approach, acknowledging the emotion while offering constructive advice. At high intensity (80\%), the persona shifts entirely to an empathetic supporter, using emotional language and focusing on validating the user's worth rather than immediate problem-solving. This confirms that Fusian effectively manipulates the underlying semantic focus of the model.

\subsection{Analysis of Personality Manifold}

\begin{figure}[t]
  \centering
  \includegraphics[width=\columnwidth]{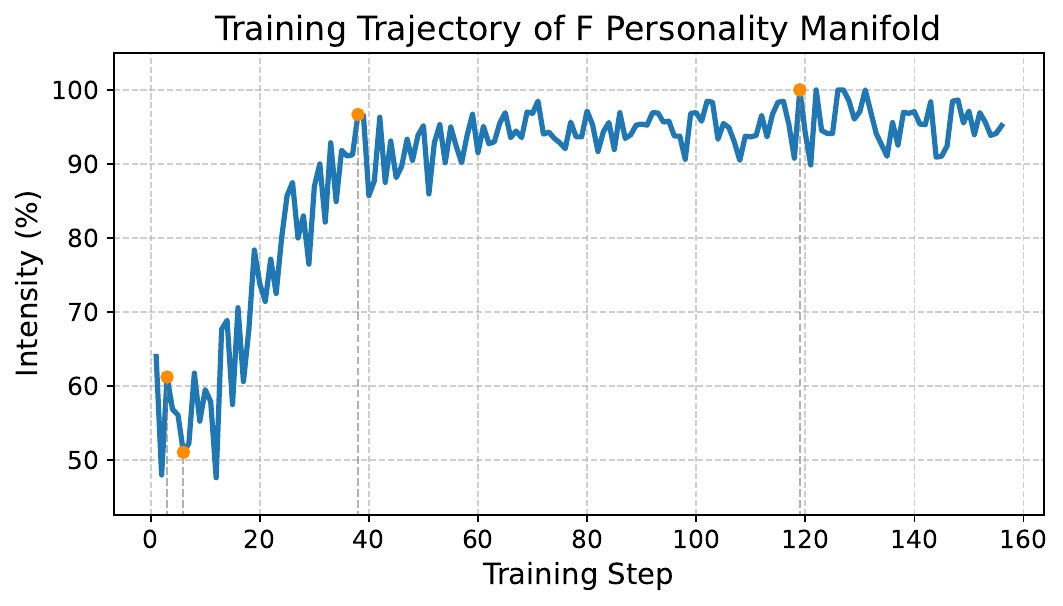}
  \caption{The personality manifold from the SFT trajectory. Trait intensity evolution reveals non-monotonicity and diminishing returns.}
  \label{fig:manifold}
\end{figure}

To gain a deeper understanding of the properties of the learned personality subspace, we analyze the training trajectory captured in Stage 1 (\autoref{fig:manifold}). The selection of LoRA adapters reveals two critical characteristics of the optimization landscape.

\paragraph{Non-Monotonicity and Local Fluctuations}
Contrary to the assumption that trait intensity increases monotonically with training steps, we observe local fluctuations in the early phases. For instance, in the "Feeling" (F) trajectory, the adapter at Step 3 exhibits a higher trait intensity ($61.22\%$) than Step 6 ($51.06\%$). This instability suggests that the model's traversal of the personality manifold is noisy initially. Consequently, our \textit{Stable Basis Selection} strategy, which filters based on variance and value rather than step count, is essential for constructing a reliable interpolation basis.

\paragraph{Diminishing Returns}
The distribution of selected checkpoints highlights a "diminishing returns" phenomenon. The model rapidly acquires the majority of the trait intensity (reaching $\sim96\%$ by Step 38) but requires a disproportionately large number of steps to perfect the final increment (extending to Step 119 for $100\%$). This indicates that the gradient of personality intensity flattens significantly near the extremes, making the "long tail" of training crucial for achieving full intensity control.

\section{Conclusion}

We introduce Fusian to address the lack of fine-grained, continuous personality control in LLMs. Departing from discrete or prompt-based strategies, Fusian leverages the latent structure of SFT trajectories and performs RL-driven Multi-LoRA fusion to model personality as a continuous manifold. Through dynamic weight computation over trajectory checkpoints, our approach enables precise and stable intensity alignment while accounting for the non-linear parameter–semantic relationship.

\newpage
\section*{Limitations}

Our current framework is designed to control a single MBTI dimension at a time. Extending the methodology to simultaneously fuse adapters from multiple, potentially interacting dimensions (e.g., Extraversion and Thinking) to compose a complete personality profile requires further exploration. Additionally, the computational cost of evaluating the model during the RL phase is non-trivial, although the training is done offline. Nevertheless, Fusian is not inherently tied to the MBTI framework and can be readily extended to other personality models, such as the Big Five, by adapting the training dataset and evaluation questionnaires to the corresponding traits. Furthermore, our experiments are currently limited to Qwen3-14B, and the effectiveness of Fusian across other open-source LLMs has not yet been systematically evaluated. Since different models may exhibit distinct parameter spaces and personality learning dynamics, future work should investigate Fusian across models with different architectures and parameter scales, which may also require model-specific trajectory collection and hyperparameter tuning.

\section*{Acknowledgements}

This study is supported by National Key R\&D Plan of China under Grant No. 2024YFB4303900. The authors would also like to thank National Supercomputer Center in Guangzhou for providing high performance computational resources.


\bibliography{ref}

\newpage
\clearpage
\appendix

\section{Additional Training Details}
\label{sec:appendix_training_details}

\paragraph{Adapter Selection}
From the trajectory library, we select $N=10$ stable adapters using a sliding window variance filter with window size $W=3$ and variance threshold $\sigma^2_{thresh}=10.0$.

\paragraph{Policy Network and Sampling}
The Policy Network outputs the concentration parameters of a Dirichlet distribution used to sample LoRA fusion weights.
A small offset $\epsilon=0.01$ is added to the Dirichlet parameters to ensure numerical stability.

\paragraph{Sampling Strategy}
In each training epoch, we sample 32 target personality values uniformly across the controllable range using evenly spaced points with small random perturbations.

\paragraph{Reward and Baseline}
The reward uses the exponential shaping function described in~\autoref{eq:reward}.
The reward scaling factor is set to $\lambda=20.0$.
A bonus reward $B=0.5$ is added when the absolute prediction error is below $\delta=0.01$.
To reduce variance, we use an exponential moving average baseline with decay rate $0.95$.

\paragraph{Optimization}
Training uses gradient accumulation with 4 steps.
Gradients are clipped to a maximum norm of $0.5$ to stabilize policy updates.

\paragraph{Entropy Regularization}
The entropy coefficient is initialized at $0.1$ and decays by a factor of $0.9$ per epoch, with a minimum value of $0.01$.
\end{document}